\documentclass[10pt,twocolumn,letterpaper]{article}

\usepackage{wacv}
\usepackage{times}
\usepackage{epsfig}
\usepackage{graphicx}
\usepackage{amsmath}
\usepackage{amssymb}
\usepackage{multirow}
\usepackage{algorithm}
\usepackage{algpseudocode}
\usepackage{makecell}
\usepackage{enumitem}

\wacvfinalcopy 

\def\wacvPaperID{992} 

\ifwacvfinal\fi
\begin{document}

\title{Fine-grained Image Classification and Retrieval by Combining Visual and Locally Pooled Textual Features}

\author{Andres Mafla ~~ Sounak Dey ~~ Ali Furkan Biten ~~ Lluis Gomez ~~ Dimosthenis Karatzas\\
Computer Vision Center, UAB, Spain\\
{\tt\small \{andres.mafla, sdey, abiten, lgomez, dimos\}@cvc.uab.es}
}

\maketitle
\ifwacvfinal\thispagestyle{empty}\fi

\begin{abstract}
Text contained in an image carries high-level semantics that can be exploited to achieve richer image understanding. In particular, the mere presence of text provides strong guiding content that should be employed to tackle a diversity of computer vision tasks such as image retrieval, fine-grained classification, and visual question answering.
In this paper, we address the problem of fine-grained classification and image retrieval by leveraging textual information along with visual cues to comprehend the existing intrinsic relation between the two modalities. The novelty of the proposed model consists of the usage of a PHOC descriptor to construct a bag of textual words along with a Fisher Vector Encoding that captures the morphology of text. This approach provides a stronger multimodal representation for this task and as our experiments demonstrate, it achieves state-of-the-art results on two different tasks, fine-grained classification and image retrieval. 
The code of this model will be publicly available at 
\footnote{http://www.github.com/DreadPiratePsyopus/Fine\_Grained\_Clf}.
\end{abstract}

\section{Introduction}
Written communication is arguably one of the most important human inventions
that allows the transmission of information in an explicit manner. Moreover, given the fact that text is omnipresent 
in man made scenarios~\cite{veit2016coco, karatzas2015icdar}, as well as the implicit relation between visual information and scene text instances, the design of holistic computer vision models for scene interpretation is fundamental.

With the purpose of designing a holistic model, in this work we leverage textual information applied to the problem of fine-grained classification and image retrieval.
Fine-grained classification tackles the problem of classifying different object instances that are visually similar and difficult to discriminate.
The complexity of this task lies in finding discriminative features 
which often require domain specific knowledge \cite{maji2013fine, xiao2015application}. 


\begin{figure}[htb]

\centering
\includegraphics[width=0.9\columnwidth]{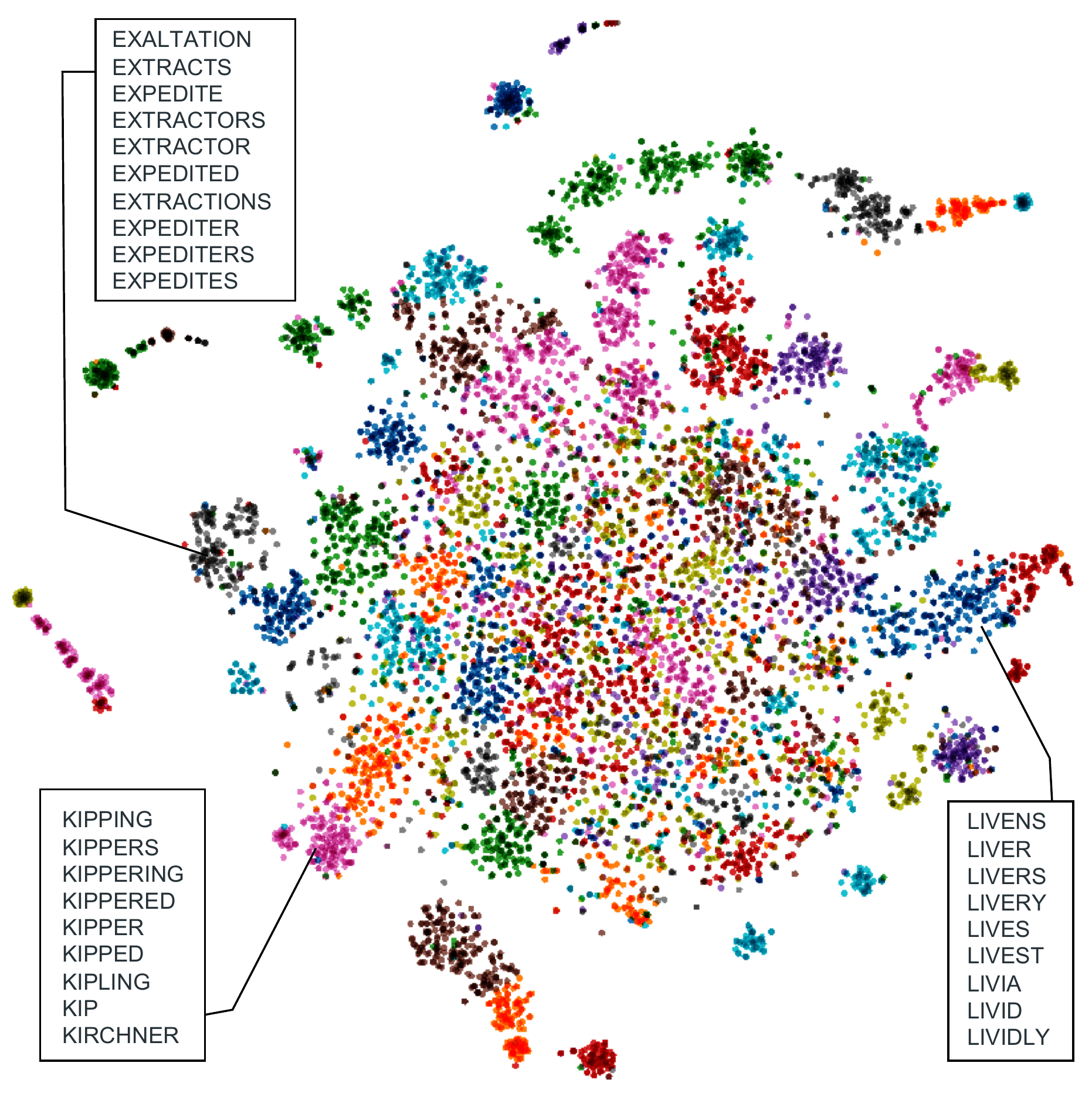}
\caption{T-SNE Visualization~\cite{maaten2008visualizing} of the $300$ dimensional PCAed PHOCs in a two dimensional space. Words with similar morphology are clustered together by a Gaussian Mixture Model, thus making such a descriptor suitable and powerful enough to discriminate text for a fine-grained classification task.}

\label{fig:gmm_cluster}
\end{figure}

An early work that demonstrated the importance of text (domain specific knowledge) for fine-grained storefront classification was put forward by Movshovitz \etal~\cite{movshovitz2015ontological}, in which the trained classifier learned automatically to attend to text found in an image as the sole way of solving the task.
Since then, there has been additional research that explicitly combines textual and visual cues, being the work presented by Karaoglu~\textit{et al.}~\cite{karaoglu2013text, karaoglu2017words} and Bai \textit{et al.}\cite{bai2018integrating} the most related ones to our paper. 
In this work, we propose the usage of a state of the art text retrieval model presented by Gomez \etal~\cite{Gomez_2018_ECCV} to detect and obtain the Pyramidal Histogram of Characters (PHOC) of scene text. We use the PHOC descriptors extracted from images and explore different fusion strategies to merge the visual and textual modalities. Additionally, we construct a Fisher Vector (FV) Encoding from the obtained PHOCs to be used as a fixed-length text feature in our pipeline and further improve the classifier results. 
Our model leverages the visual features combined with the morphology of a word (refer to Figure~\ref{fig:gmm_cluster}), 
that belong to specific fine-grained classes, without the need to understand them semantically. Contrary to previous methods, this approach is especially useful when dealing with text recognition errors and named entities which are often difficult to encode in a purely semantic space. The combination of these two modalities produce an output probability vector that addresses the classification task at hand. As an additional application, we evaluate the proposed model on fine-grained image retrieval in available datasets.  
Overall, the main contributions of our work are:
\begin{itemize}[noitemsep,nolistsep]
    \item We propose a novel architecture that achieves state of the art on fine-grained classification by considering text and visual features of an image.
    \item We show that by using Fisher Vectors obtained from PHOCs of scene text, we obtain a more robust representation in which 
    words with similar structure get encoded on the same Gaussian component, thus creating a more powerful discriminative descriptor than PHOCs alone.
    \item We provide exhaustive experiments in which we compare the performance of different alternative modules in our model and previous state of the art.
\end{itemize}

\section{Related Work}
\label{sec:Related_work}
\subsection{Scene Text Detection and Recognition}
Even though deep learning has made significant progress~\cite{lecun2015deep}, localizing and recognizing text in images still remains an open problem in the computer vision community due to the ample variety of text occurrences in natural images~\cite{zhu2016scene}. Essentially a system capable of reading text requires two steps, detection and recognition. Jaderberg \textit{et al.} \cite{jaderberg2016reading} tackles this problem by generating text proposals that were refined by a CNN. The bounding boxes obtained were used as input to another CNN that was trained to classified them according to a fixed dictionary. In another work,~\cite{gupta2016synthetic} defined a Fully Convolutional Regression Network to detect text by regressing bounding boxes and the same classification network as~\cite{jaderberg2016reading} was employed for text recognition. 
More recent approaches use customized variations of object detectors fine-tuned to detect text instances such as~\cite{kim2016pvanet} and~\cite{liu2016ssd} resulting in models proposed by \cite{zhou2017east} and~\cite{liao2017textboxes, liao2018textboxes++}. 
Recently, the community attention has placed an additional effort in the development of end-to-end models. 
The main existing notion is that features that help to improve detection are also useful at the moment of recognizing text instances. He \textit{et al.} \cite{he2018end} uses a CNN to extract proposals, which are fed into an LSTM (Long-Short Term Memory) to refine the bounding boxes that are later employed as input to yet another LSTM to perform recognition. 
In parallel, additional work has been conducted into the development of multilingual scene text recognizers, such as the work of \cite{buvsta2018e2e} which consists on two CNNs. The first one is optimized to detect text and a the second one employs a Connectionist Temporal Classification (CTC)~\cite{graves2006connectionist} module for recognition, while training both in an end to end manner.

In this work, we leverage the Pyramidal Histogram Of Characters (PHOC) descriptor~\cite{almazan2014word, sudholt2017learning} (see Figure~\ref{fig:phoc_sample}) commonly used to query a given text instance in handwritten documents and natural scene images. The PHOC of a word encodes the position of a specific character in a particular spatial region of the detected text instance. 
Such a descriptor has proven to perform as the state of the art in scene text retrieval~\cite{Gomez_2018_ECCV}, and as our experiments show, encoding it with the Fisher Vector~\cite{perronnin2007fisher} provides an improved text descriptor for fine-grained classification. 

\begin{figure*}[ht]
\begin{center}
\includegraphics[width=0.95\linewidth]{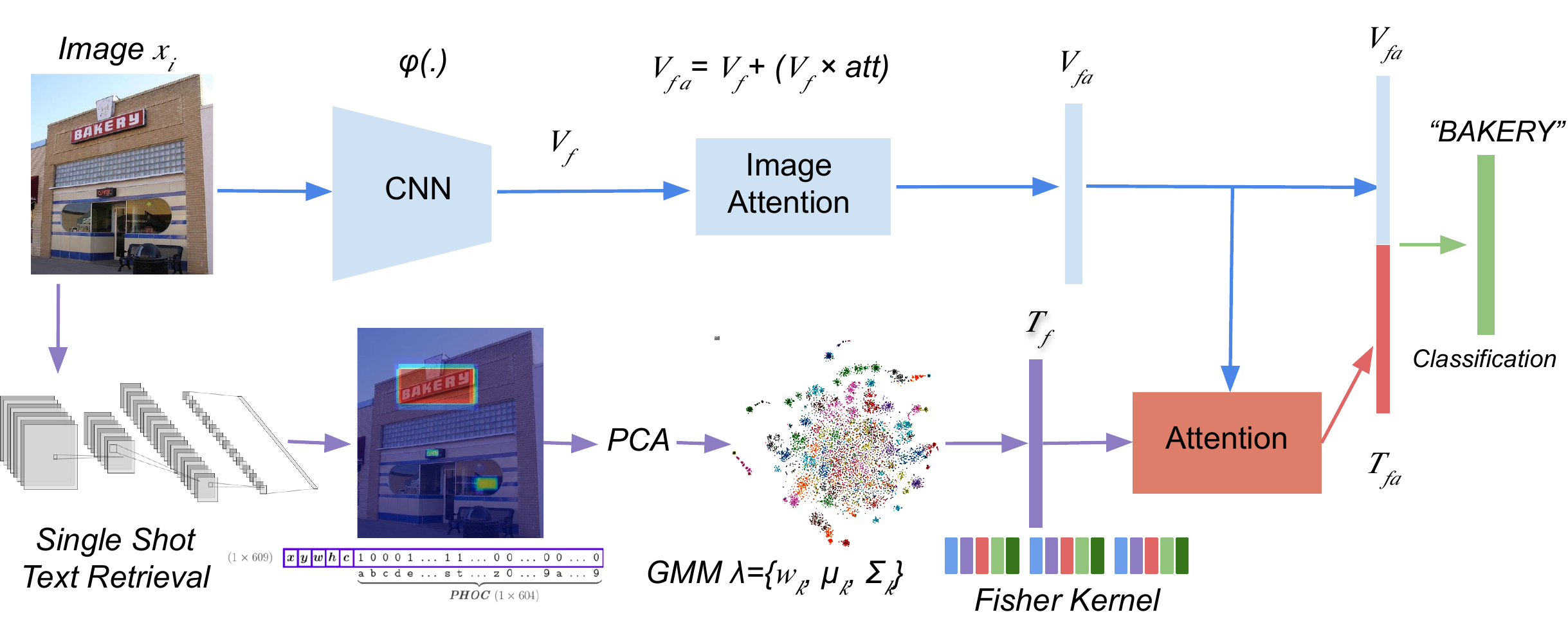}
\end{center}
\caption{Proposed model pipeline. The PHOCs obtained from~\cite{Gomez_2018_ECCV} are used to compute a Fisher Vector that yields a compact morphology based descriptor suitable to discriminate features from visually similar objects.}
\label{fig:model}
\end{figure*}

\subsection{Fine-Grained Classification}
Recent works on fine-grained classification base their approach on localizing salient parts of an image \cite{deng2013fine,yang2012unsupervised}, and use the saliency maps to classify the objects. 
Later approaches such as the one of Tang \etal~\cite{tang2017learning}, use a weakly supervised method to find discriminative features and leverage them to perform the classification between similar instances. Other methods use existing prior knowledge from unstructured text to propose a semantic embedding that differentiates similar classes~\cite{xu2018fine}. A self-supervision method is introduced in~\cite{yang2018learning} that learns to propose significant image regions to find inter-class discriminative features.

More related to our work,~\cite{karaoglu2017words} tackles this task by extracting visual features with a pre-trained GoogleNet \cite{szegedy2015going} and a Bag of Words feature to represent the text instances found in an image and further classify them. More recently, Bai \etal~\cite{bai2018integrating} use a similar approach and extract visual features using a GoogleNet and a combination of two models: \cite{liao2017textboxes} to detect and \cite{shi2017end} to recognize text. The text found is represented as GloVe features \cite{pennington2014glove}, a word embedding that is further used with attention on the visual features to find a semantic relation between the two modalities to classify the image.


\subsection{Multimodal Fusion}
\label{sec:mult_fusion}

The combination of different modalities provides a richer content description rather than one modality alone, therefore the contained knowledge should be leveraged to further exploit explicit information according to the task \cite{srivastava2012multimodal}. 
In this work we explore other fusion methods used in multimodal learning, that shows a performance increase especially in tasks that require exploiting two modalities such as Visual Question Answering (VQA) and Visual Relationship Detection (VRD). 

One of the initial works presented by~\cite{ben2017mutan}, modeled a Tucker decomposition of the bilinear interaction of two distinct modalities. Later, a Multimodal Low-rank Bilinear Attention Network (MLB) was proposed by~\cite{kim2016hadamard}, in which the result of the fusion of two modalities was based on a low-rank bilinear pooling operation using the Hadamard product along with an attention mechanism. 
A factorized bilinear pooling (MFB) is proposed by~\cite{yu2017multi}, where each third mode section of the tensor is constrained by a rank. Later methods, such as a Multimodal Factorized High-order pooling (MFH) fusion was presented by~\cite{yu2018beyond}, which uses a high-order fusion formed by cascaded MFB modules. In the work conducted by~\cite{ben2017mutan}, a bilinear pooling is performed where the tensor is represented as a Tucker decomposition. The obtained main tensor has the same rank constrain as the MFB technique.  Lately, a Multimodal Bilinear Superdiagonal Block (Block) fusion strategy based on the work presented by~\cite{ben2019block}, has achieved state of the art results in VQA and VRD. 

\section{Proposed Model}
\label{sec:Proposed_Model}

The devised model consists mainly in four processing blocks: visual features extraction, textual features extraction, attention unit and classification. The whole model pipeline is shown in Figure \ref{fig:model}.

The first block extracts the visual features from a given image and produces a fixed size representation of it. The second block consists of extracting the PHOC representation of each text instance found in an image and use a pre-trained Gaussian Mixture Model (GMM) to obtain the correspondent FV descriptor. The third block consists of an attention unit that multiplies learned weights with the encoded FV depending on the visual features extracted previously. Finally, the last block consists of a concatenation of the two different modalities followed by a fully connected layer to obtain a probability output vector which is used for classification. For the rest of the paper, let $\mathcal{C}$ be the set of all possible categories in a given dataset; $\mathcal{X}=\{x_i \}_{i}^N$ be the set of images; $l_x:\mathcal{X}\rightarrow\mathcal{C}$ be the labelling function.
\subsection{Visual Features}

In our model, we use a Convolutional Neural Network (CNN)~\cite{he2016deep} pre-trained on ImageNet~\cite{deng2009imagenet} as a visual feature extractor, denoted as $\phi(\cdot)$. We use the output of the last convolutional block of $\phi(\cdot)$ before the last average pooling layer as the visual features, denoted as $V_f$.
Attention on visual features has proven to yield improved performance on several tasks. As it is presented by~\cite{dey2019doodle}, we compute a soft-attention mechanism due to its differentiable properties, thus allowing an end-to-end learning. The proposed attention function learns an attention mask $att$ which assigns weights to different regions of an image given a feature map $V_f$. The attention mask is learned by applying $1\times1$ convolution layers on the output features from the CNN. Lastly, to obtain the final output of the attention module along with the visual features, the operation is computed by $V_{fa} = V_f + (V_f \times att)$. 

\subsection{Textual Features}
\label{sec:textual_features}
Methods shown in previous works~\cite{karaoglu2017words,bai2018integrating} contain mainly three drawbacks. First, the employed text recognizers are bound to a fixed dictionary, which may or may not include the exact words that are present in the image. Second, some words that are contained in the fixed recognition dictionary may not exist in the proposed semantic embedding (GloVe, Word2Vec) such as license plates, brand names, acronyms, etc. 
Third, any mistake committed by the recognizer will yield a vector embedding that lies far from the semantic embedding of the correct word. Contrary, 
correct recognition of semantically similar words that might indicate different fine-grained classes will lead to embeddings close to each other, which are not discriminative enough to perform correct classification. This is the case of similar semantic words such as restaurant and steakhouse, cafe and bistro, coke and pepsi among some other sample classes from the datasets used.

\begin{figure}[h]
\begin{center}
\includegraphics[width=\linewidth]{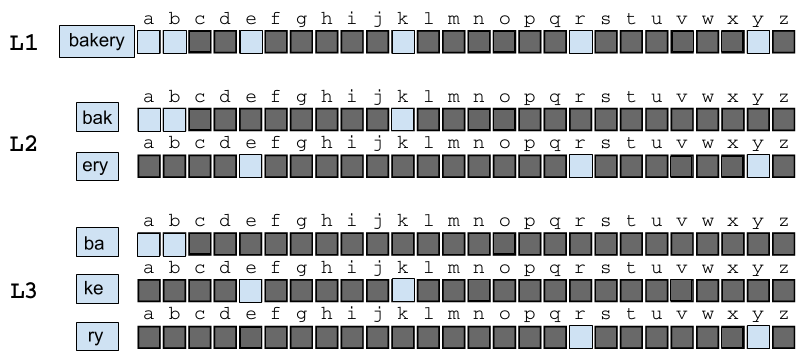}
\end{center}
\caption{Levels $1$ to $3$ of the PHOC of the word "bakery". The final compact representation is a concatenation of the histograms of each level. Blues represent "$1$" while blacks represent "$0$". Best viewed in color.}
\label{fig:phoc_sample}
\end{figure}

In order to exploit the morphology of a word to obtain discerning features, we employ the PHOC representation. The PHOC representation employed in this work is composed by the concatenation of vectors from the levels $2$ to $5$ plus the $50$ most common bi-grams in English language. This yields a 604-dimensional discrete binary vector that represents the characters contained in a word (see Figure~\ref{fig:phoc_sample}). A dictionary given by~\cite{jaderberg2016reading} is employed to obtain a PHOC per word, in this way, we populate a matrix of this compact representation. In order to reduce the dimensionality and to find linearly uncorrelated variables of this compact vector, a Principal Component Analysis (PCA) is performed. This procedure yields a more compact but at the same time informative vectorial representation of a given word. 

The obtained data points were used to construct a Gaussian Mixture Model (GMM) \cite{gregor1969algorithm} formed by $K$ Gaussian components. We denote the parameters of the $K$-component GMM by $\lambda=\left\{w_{k}, \mu_{k}, \Sigma_{k}, k=1, \ldots, K\right\}$, where $w_k, \mu_k$ and $\Sigma_k$ are respectively the mixture weight, mean vector and covariance matrix of Gausssian $k$. We define:
\begin{equation}
\label{e:u_lambda}
u_{\lambda}(x)=\sum_{k=1}^{K} w_{k} u_{k}(x)
\end{equation}
where $u_k$ denotes Gaussian $k$:
\begin{equation}
\label{e:gauss_k}
u_{k}(x)=\frac{1}{(2 \pi)^{D / 2}\left|\Sigma_{k}\right|^{1 / 2}} \exp \left\{-\frac{1}{2}\left(x-\mu_{k}\right)^{\prime} \Sigma_{k}^{-1}\left(x-\mu_{k}\right)\right\}
\end{equation}
and we require:
\begin{equation}
\label{e:for_all}
\forall_{k} : w_{k} \geq 0, \quad \sum_{k=1}^{K} w_{k}=1
\end{equation}

Once the GMM model is trained, it will be used to extract a single Fisher Vector representation per image which encodes its contained textual information. The textual features per image are obtained by using the model from~\cite{Gomez_2018_ECCV}. Given an input image, the model outputs a list of $\mathcal{B}$ bounding boxes, each one containing a confidence score $\mathbb{C}$ and a PHOC prediction. 

 
We get the top-$m$ object proposals set $\mathcal{O}_m:=\left\{o \in \mathbb{C}_{i}: o \geq c, \forall c \in \mathbb{C}_{i}\right\}$. The resulting PHOCs $\in[0,1]^{d \times N}$, where $d$ is the dimensionality of the PHOC embedding obtained and $N$ the recognized words embedded in the PHOC space. It is essential to note that the model from~\cite{Gomez_2018_ECCV} is able to generalize and construct PHOCs from previously unseen samples, out of vocabulary words and different languages that employ s similar character set (e.g. Latin), making it suitable for the task at hand. 
Afterwards, we project each embedded textual instance of the obtained descriptors into a reduced dimensional space by employing PCA. The resulting vectors are used to obtain the Fisher Vector \cite{perronnin2007fisher} from the previously trained GMM. The GMM associates each PCAed vector $o_i$ to a component $k$ in the mixture model with a weight given by the posterior probability:
\begin{equation}
\label{e:post_prob}
q_{i k}=\frac{\exp \left[-\frac{1}{2}\left(o_{i}-\mu_{k}\right)^{T} \Sigma_{k}^{-1}\left(o_{i}-\mu_{k}\right)\right]}{\sum_{t=1}^{K} \exp \left[-\frac{1}{2}\left(o_{i}-\mu_{t}\right)^{T} \Sigma_{k}^{-1}\left(o_{i}-\mu_{t}\right)\right]}
\end{equation}
For each mode $k$, consider the mean and the covariance deviation vectors
\begin{equation}
\label{e:cov_dev_vec}
\begin{split}
u_{j k}=\frac{1}{N \sqrt{w_{k}}} \sum_{i=1}^{N} q_{i k} \frac{o_{j i}-\mu_{j k}}{\sigma_{j k}},
\\
v_{j k}=\frac{1}{N \sqrt{2 w_{k}}} \sum_{i=1}^{N} q_{i k}\left[\left(\frac{o_{j i}-\mu_{j k}}{\sigma_{j k}}\right)^{2}-1\right]
\end{split}
\end{equation}
where $j=1, 2, \dots, D$ spans the vector dimensions. The FV of a given image $I$ is simply the concatenation of the obtained vectors $u_k$ and $v_k$ for each of the $K$ components in the Gaussian mixture model.
\begin{equation}
\label{e:cov_dev_vec2}
T_f=\begin{bmatrix}%
\ {\cdots} & \ {\mathbf{u}_{k}} & \ {\cdots} & \ {\mathbf{v}_{k}} & \ {\cdots} \
\end{bmatrix} ^\intercal
\end{equation}
The FV and the GMM encode inherently similar information. This takes place because they both include statistics of order $0$, $1$ and $2$~\cite{sanchez2013image, perronnin2007fisher}. However, the FV provides a vectorial representation which is more compact, faster to compute and suitable for processing.
The dimension of the FV obtained, noted as $T_{f}$, is given by $(2 \times d \times K)$, where $d$ is the PHOC dimension after performing the PCA and $K$ is the number of Gaussian clusters. The intuition captured by the FV is to compute the gradient of a PHOC sample (bag of textual features) that shows the probability of belonging to each of the Gaussian components, which can be understood as a probabilistic textual vocabulary based on its morphological structure (see Figure~\ref{fig:gmm_cluster}).


\subsection{Attention on features}
In the proposed fine-grained classification task we can intuitively state that there will be some recognized text that is more relevant than others at the moment of discriminating similar classes. 
Therefore, it is important to capture the inner correlation between the textual and visual features. 
To adhere this idea into our pipeline, we propose a modified attention mechanism inspired from \cite{you2016image}. The attention mechanism learns a tensor of weights $W$ that is used between the visual features and the obtained FV. The implemented attention is defined by:

\begin{equation}
W_a = Softmax( tanh (V_{fa}^{T}\cdot W \cdot T_{f}))
\end{equation}
\begin{equation}
T_{fa} = W_{a} \cdot T_{f}
\end{equation}
The resulting tensor $W_a$, contains a normalized attention vector that is multiplied with the textual features $T_f$ to obtain the final attended textual features $T_{fa}$.

The obtained attended textual features $T_{fa}$ and the visual features $V_{fa}$ are concatenated, such that the final features are formed by $F = [V_{fa}, T_{fa}]$. Finally, the resulting vector serves as input to a final classification layer that outputs the probability of a given class.
The proposed network is trained to optimize the cross entropy loss function
given by:

\begin{equation}
\label{e:loss_function}
J(\theta) = -\frac{1}{N}\sum_{n=1}^{N}\sum_{i=1}^{\mathcal{C}}l_i^{n} log (\hat{l}_i^{n})
\end{equation}

\section{Experiments and Results}
\label{sec:Experiments}
The following section describes the datasets employed, the implementation details along with the analysis of the results obtained from the experiments conducted. 

\subsection{Datasets}
\subsubsection{Con-Text Dataset} Originally presented by~\cite{karaoglu2013text}, is a dataset taken from the ImageNet~\cite{deng2009imagenet} "building" and "place of business" sub-categories. It consists of $28$ categories with $24,255$ images in total. The classes from this dataset are visually similar (Pizzeria, Restaurant, Dinner, Cafe) and requires text to successfully perform a fine-grained classification. The dataset was not built for text recognition purposes, thus not all images contain text in them. A high variability of text size, location and font styles make text recognition on this dataset a challenging task.
\subsubsection{Drink Bottle Dataset} Dataset presented by~\cite{bai2018integrating}
comprises the sub-categories soft drink and alcoholic drink found on ImageNet\cite{deng2009imagenet}. There are 18,488 images divided in 20 categories. The dataset contains several not common, occluded, rotated, low quality and blurred text instances which increases the difficulty of performing successful text recognition.

\subsection{Implementation Details}
The visual features of the proposed model are taken by attending the features of the output of the last block layer of the Resnet$152$ before the last average pooling layer. These features are passed through a fully connected layer to down-sample them to a final dimension of $1\times1024$. To construct the textual features, a maximum number of $N_{max} = 15$ PHOC proposals are obtained per image. If a lesser number of PHOC proposals are obtained, a zero padding scheme is employed to fix the size of the input features. The resulting PHOCs are reduced in size through PCA, to obtain features of a dimensionality of $N_{max}\times300$. \\
The Fisher Vector is calculated from the PCA-ed PHOCs by employing a pre-trained Gaussian Mixture Model as it is described in Section~\ref{sec:textual_features}. The trained GMM employs $64$ Gaussian components thus yielding a FV of $1\times38400$ dimension. The obtained textual features are down-sampled by passing them through a fully connected layer to finally obtain a resulting size of $1\times512$ before the attention mechanism 
is computed. The attention between both modalities produces an output vector of $1\times512$, that multiplies the learned weights to the textual features. As the last step, a concatenated vector of the visual and textual features ($dim = 1\times1536$) is used to produce the final classification probability vector.

The network is trained for $30$ epochs with the combination of RAdam~\cite{liu2019radam} and the Lookahead~\cite{zhang2019lookahead} optimizers. The batch size employed in all our experiments is $64$, with a learning rate of $0.001$, momentum of $0.9$ that decays by $0.1$ every $10$ epochs.

\subsection{Comparison with the State of the Art}
When comparing our method to the current state of the art, it is evident that the proposed pipeline consistently outperforms previous approaches. The performance of our method is shown in Table~\ref{table:context_bottles_short} (see the Supplementary Material Section for the results of each of the classes found in the Con-Text and Drink Bottle datasets respectively). As it can be seen, our method surpasses~\cite{bai2018integrating} in the Drink Bottle dataset by a significant margin, however this margin is smaller in the Con-Text dataset. Nonetheless, it is important to note that the method presented by~\cite{bai2018integrating} employs two additional classifiers to solve this task, thus relying on an ensemble model. Such kind of adopted approaches require longer training times, as well as more computation resources since several deep networks need to be trained. Therefore, when comparing to the single classifier presented by~\cite{bai2018integrating}, our model offers a significant improvement. In the upcoming sections, we provide explanations and exhaustive experimentation that shows the main strengths and advantages of our model.
\begin{table}[h]
\normalsize
\setlength\tabcolsep{2pt}
\begin{center}
\begin{tabular}{l|l|l}
\hline
\textbf{Method} & \textbf{Con-Text} & \textbf{Bottles}\\ \hline
Karaoglu\cite{karaoglu2013text} & $39.0$ & $-$\\ \hline
Karaoglu\cite{karaoglu2017words} & $77.3$ & $-$\\ \hline
Bai\cite{bai2018integrating} & $78.9$ & $-$\\ \hline
Bai*\cite{bai2018integrating} & $79.6$ & $72.8$\\ \hline
\textbf{Ours} & \boldmath$80.2$  & \boldmath$77.4$\\ \hline
\end{tabular}
\end{center}
\caption{Classification performance for two state-of-the art methods and our proposed model on the Con-Text and Bottles dataset. The results presented by~\cite{bai2018integrating} depicted with * are based on an ensemble model.}
\label{table:context_bottles_short}
\end{table}

\subsection{Importance of Textual Features}
Several baselines of growing complexity were defined in order to: assess the effectiveness of the proposed model, discern the added performance of employing textual features along visual ones and to verify the improvement obtained from using a fusion mechanism.\\
\textbf{Visual Only:} This baseline assesses the performance of the CNN encoder based on visual features solely. To this end, the 2048 dimensional output features $V_f$, serve as the input to a fully connected layer according to the number of classes of the evaluated dataset.\\
\textbf{Textual Only:} We evaluate the performance of two state of the art text recognizers: Textspotter~\cite{he2018end} and E2E\_MLT~\cite{buvsta2018e2e} along with the most confident PHOCs obtained from the model presented by~\cite{Gomez_2018_ECCV}. 
\begin{figure}[ht]
\begin{center}
\begin{tabular}{p{0.45\linewidth} p{0.45\linewidth}}
    \includegraphics[width=0.95\linewidth,height=0.8\linewidth]{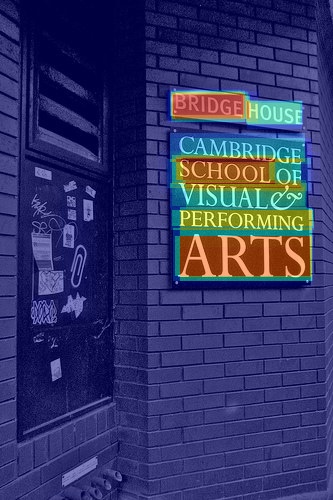}  &
    \includegraphics[width=0.95\linewidth,height=0.8\linewidth]{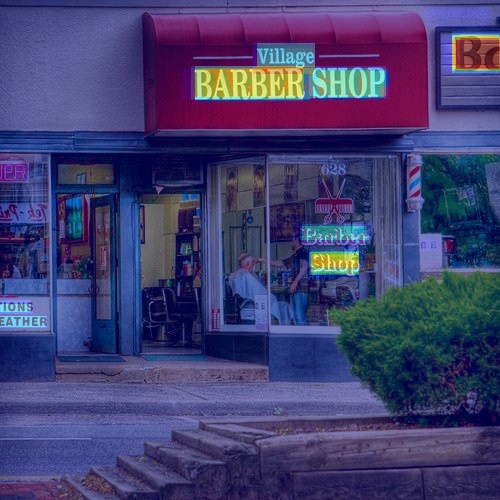} 
    \\
    & \\
    \includegraphics[width=0.95\linewidth,height=0.8\linewidth]{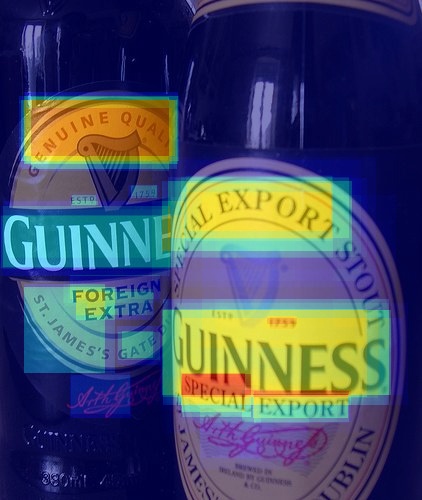} 
    & 
    \includegraphics[width=0.95\linewidth,height=0.8\linewidth]{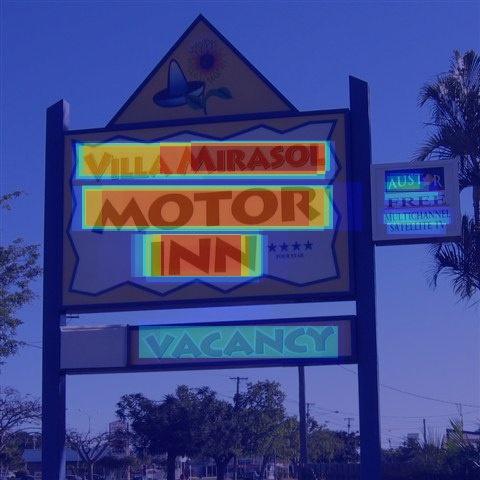}
\end{tabular}
\end{center}
\caption{Heat maps obtained according to the confidence detection score of the predicted PHOCs.}
\label{fig:heatmaps}
\end{figure}
For illustration purposes, Figure~\ref{fig:heatmaps} shows heat maps obtained by employing the model from~\cite{Gomez_2018_ECCV} according to the confidence scores obtained when a text instance is detected. It is important to note that Textspotter~\cite{he2018end} is bound to a dictionary to output the final recognized word, whereas the multilingual model E2E\_MLT  from~\cite{buvsta2018e2e} is not.
The recognized text is embedded with pretrained versions of GloVe~\cite{pennington2014glove}, FastText~\cite{bojanowski2017enriching} and Word2Vec~\cite{mikolov2013distributed}, finally outputting tensors of size $N_{max} \times 300$, which in our experiments $N_{max}= 15$. When working with PHOCs, the output vector has a size $N_{max} \times 604$.
As we can observe in Table~\ref{tab:results1}, in the visual only baseline, the ResNet152 CNN~\cite{he2016deep} performed better in this task, due to the major expressiveness of the model and the residual block architecture that it is based on. 

\begin{table}[H]
\begin{center}
\small
\begin{tabular}{c|l|l|l}
\multicolumn{1}{l|}{}       & \textbf{Model}         & \textbf{Con-Text} & \textbf{Bottles} \\ \hline
\multirow{2}{*}{\textbf{Visual}} & GoogLenet              & $61.21$             & $64.93$            \\
                            & Resnet-152             & $63.70$              & $66.56$            \\ \hline
\multirow{7}{*}{\textbf{Textual}} & Texspotter+w2v         & $35.09$             & $50.68$            \\
                            & Texspotter+glove       & $34.52$             & $50.26$            \\
                            & Texspotter+fasttext    & $36.71$              & $51.93$             \\
                            & E2E\_MLT+w2v           & $44.36$             & $43.98$            \\
                            & E2E\_MLT+glove         & $44.25$             & $42.64$            \\
                            & E2E\_MLT+fasttext      & $45.07$             & $44.31$            \\
                            & \textbf{PHOC}          & \boldmath$49.18$    & \boldmath$52.39$   \\ \cline{2-4} 
\multicolumn{1}{l|}{}       & \textbf{Fisher Vector (PHOC)} & \boldmath$63.93$    & \boldmath$62.41$ \\ \hline
\end{tabular}
\end{center}
\caption{Visual only and Textual only results. The textual only results were performed on the subset of images that contained spotted text. The metric depicted is the mean Average Precision (mAP in \%).}
\label{tab:results1}
\end{table}

In the text only baseline, by using standard text recognizers we can observe that the E2E\_MLT performs better in the Con-Text dataset, whereas the Textspotter model surpasses E2E\_MLT in the Drink Bottle dataset. Nonetheless, both of them are outperformed by employing the PHOCs obtained from~\cite{Gomez_2018_ECCV} as the word embedding. This effect is due to the inherent morphological nature of the PHOC embedding.
\setlength\tabcolsep{7pt}
\begin{table*}[!tp]
\small
\normalsize
\begin{center}
\begin{tabular}{c|l|l|l|l|l|l|l|l|l|l}
\multicolumn{1}{l|}{} & \textbf{Fusion} & \textbf{T+W} & \textbf{T+G} & \textbf{T+F} & \textbf{E+W} & \textbf{E+G} & \textbf{E+F} & \textbf{PHOC} & \textbf{FV(F)} & \textbf{FV(P)} \\ \hline
 \parbox[t]{4mm}{\multirow{6}{*}{\rotatebox[origin=c]{90}{\textbf{Con-Text}}}} & \textbf{Concat} & $73.84$        & $74.11$        & $74.33$        & $77.04$        & $77.58$       & $77.77$        & $77.45$         & $77.31$          & \boldmath$80.21^\dagger$ \\
&\textbf{Block~\cite{ben2019block}}  & $73.12$        & $73.86$        & $73.18$        & $76.97$        & $78.34$        & $78.34$        & $77.96$         & $77.87$          & $79.27$          \\
&\textbf{Mutan~\cite{ben2017mutan}}    & $72.46$        & $72.08$        & $73.47$        & $77.67$        & $77.26$        & $78.05$        & $76.97$         & $76.01$          & $78.51$         \\
&\textbf{MLB~\cite{kim2016hadamard}}    & $73.17$        & $72.18$        & $74.09$        & $77.45$        & $76.28$        & $78.81$        & $76.96$         & $76.46$          & $78.49$          \\
&\textbf{MFB~\cite{yu2017multi}}    & $73.62$        & $73.23$        & $74.42$        & $77.68$        & $76.79$        & $78.55$        & $77.56$         & $76.27$          & $78.03$          \\
&\textbf{MFH~\cite{yu2018beyond}}    & $72.95$        & $72.43$        & $74.48$        & $77.3$        & $76.64$        & $78.23$        & $77.42$         & $76.39$          & $77.58$\\
\hline
 \parbox[t]{4mm}{\multirow{6}{*}{\rotatebox[origin=c]{90}{\textbf{Drink Bottle}}}} & \textbf{Concat} & $75.05$        & $75.12$       & $75.25$        & $74.62$        & $74.91$        & $75.4$         & $75.93$         & $75.15$          & \boldmath$77.38^\dagger$ \\
&\textbf{Block~\cite{ben2019block}}  & $75.18$        & $75.31$        & $75.39$       & $74.17$       & $74.87$        & $74.94$        & $75.91$         & $75.11$          & $76.23$          \\
&\textbf{Mutan~\cite{ben2017mutan}}    & $74.48$        & $73.91$        & $74.72$        & $73.62$        & $75.12$        & $76.05$        & $75.95$         & $74.48$          & $75.97$          \\
&\textbf{MLB~\cite{kim2016hadamard}}    & $74.34$        & $73.02$        & $75.54$        & $73.55$        & $75.42$        & $75.19$        & $76.37$         & $75.07$          & $76.18$          \\
&\textbf{MFB~\cite{yu2017multi}}    & $74.25$        & $74.25$        & $75.21$        & $74.23$        & $74.88$        & $75.84$        & $76.21$         & $74.78$          & $76.01$          \\
&\textbf{MFH~\cite{yu2018beyond}}    & $73.99$        & $73.61$        & $75.36$        & $74.77$        & $75.26$        & $75.72$        & $75.98$         & $74.56$          & $75.85$ \\
\hline
\end{tabular}
\end{center}
\caption{ Results obtained by employing different fusion strategies on both the Con-Text and Drink Bottle dataset. For presentation purposes acronyms are used to represent each combination of text recognizers (Textspotter (T), E2E\_MLT (E), PHOC (P)) and word embeddings (Word2Vec (W), GloVe (G), FastText (F), Fisher Vector (FV)). The $^\dagger$ refers to the proposed model.}
\label{tab:quantitative_res}
\end{table*}

Overall, the best results in the textual only baseline are obtained by the Fisher Vector obtained from the PHOCs.
Qualitatively shown in Figure~\ref{fig:gmm_cluster}, the Gaussian Mixture gracefully captures the morphology of words obtained from PHOCs. Therefore, words with similar syntax are clustered together in the GMM, thus allowing the Fisher Vector to be a powerful descriptor relevant for this task that yields even more discriminative features than other embeddings. It is important to note as well that in our experiments, FastText performs better than Word2Vec or GloVe because it can produce embeddings of out of vocabulary words while considering word n-grams 
which strengthens our conjecture on the importance of morphology of text to solve this task. 

\subsection{Comparison of Models}
Extensive experiments were conducted regarding the different combinations of text recognizers, word embeddings and fusion techniques. Table~\ref{tab:quantitative_res} show the results obtained in both the Con-Text and Drink Bottle dataset.

When introducing fusion techniques to the models, traditional text recognizers such as E2E\_MLT performs better in Con-Text compared to Textspotter, thus achieving a higher mAP. The opposite effect is found in the Drink Bottle dataset, in which Textspotter behaves better than its E2E\_MLT. It is interesting to note that the PHOCs obtained perform consistently in both datasets, yielding comparable results to the traditional recognizers employed.
Regarding the embedding mechanism utilized, morphological embeddings (FastText, PHOC) work better than purely semantic embeddings due to the discriminative space learned. 

We can observe that the usage of fusion techniques usually improve the mAP performance obtained on each method aside from the cases when the models employ Fisher Vector features. Nonetheless, in our experiments we have not found a specific fusion technique that can be generalized for every tested method. Each fusion technique increases the performance for a specific model, being MFH and Block slightly more consistent than others. 
It is necessary to indicate that employing Fisher Vector features obtained from PHOCs consistently achieves the best performance in a general and consistent manner across both datasets.

In order to asses the efficacy of using the Fisher Vector along with another embedding that captures out of vocabulary words while at the same time considering the character morphology, we employ the Fisher Vector obtained from FastText. To this end, FastText employs character n-grams to construct a relevant vectorial representation of a word, thus it also uses syntax of a detected word. The results of the conducted experiments using Fisher Vector features from FastText and PHOC are shown in the last two columns of Tables~\ref{tab:quantitative_res}. 
There are two results to highlight obtained from this experiment. Firstly, working with PHOCs along FVs always yield better performance compared to Fasttext. The cause might be the information captured by Fasttext encapsulates morphology in the form of character n-grams, as well as semantics. Whereas the PHOC is a compact representation based solely on word morphology. 
\setlength\tabcolsep{1.2pt}
\begin{figure*}[!tp]
\begin{center}
\begin{tabular}{p{0.12\linewidth} p{0.12\linewidth} p{0.12\linewidth} p{0.12\linewidth}p{0.12\linewidth} p{0.12\linewidth} p{0.12\linewidth} p{0.12\linewidth}}
    \includegraphics[width=\linewidth,height=0.75\linewidth]{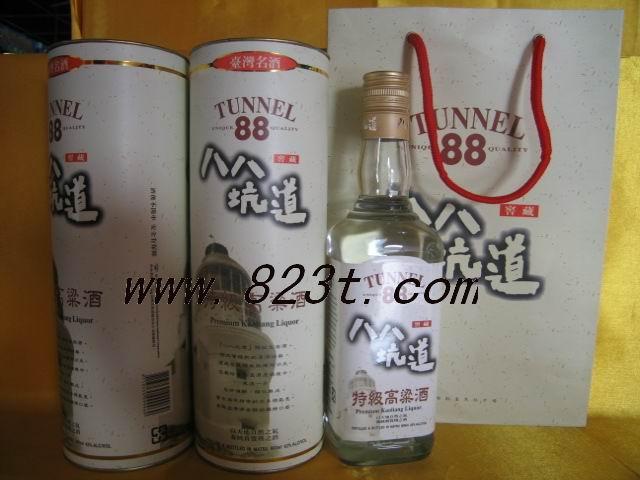} &
    \includegraphics[width=\linewidth,height=0.75\linewidth]{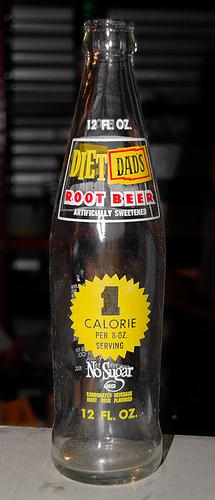} &
    \includegraphics[width=\linewidth,height=0.75\linewidth]{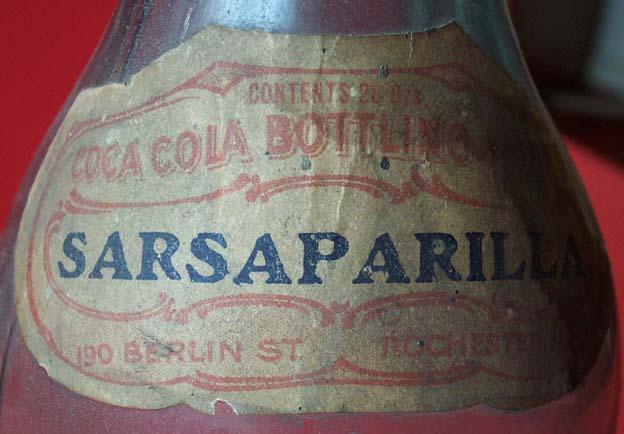} &
    \includegraphics[width=\linewidth,height=0.75\linewidth]{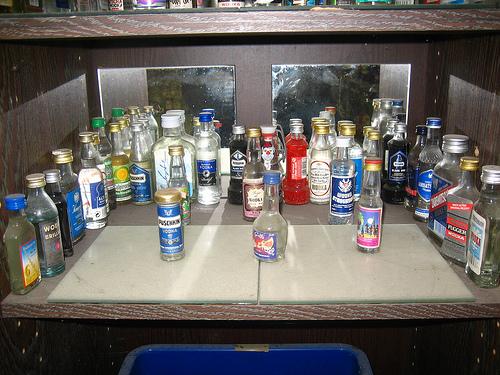} &
    \includegraphics[width=\linewidth,height=0.75\linewidth]{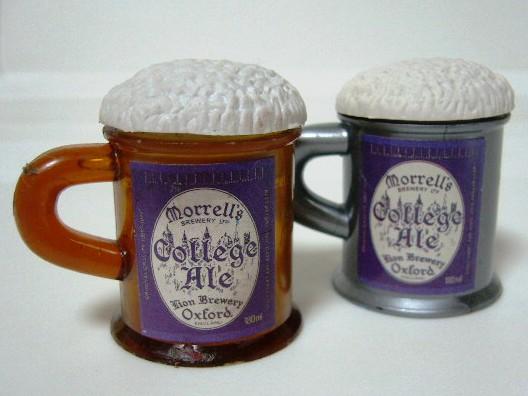} &
    \includegraphics[width=\linewidth,height=0.75\linewidth]{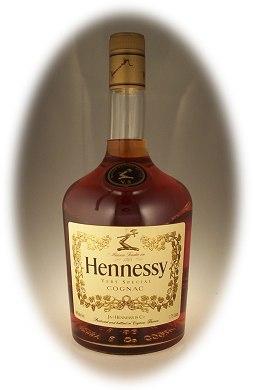} &
    \includegraphics[width=\linewidth,height=0.75\linewidth]{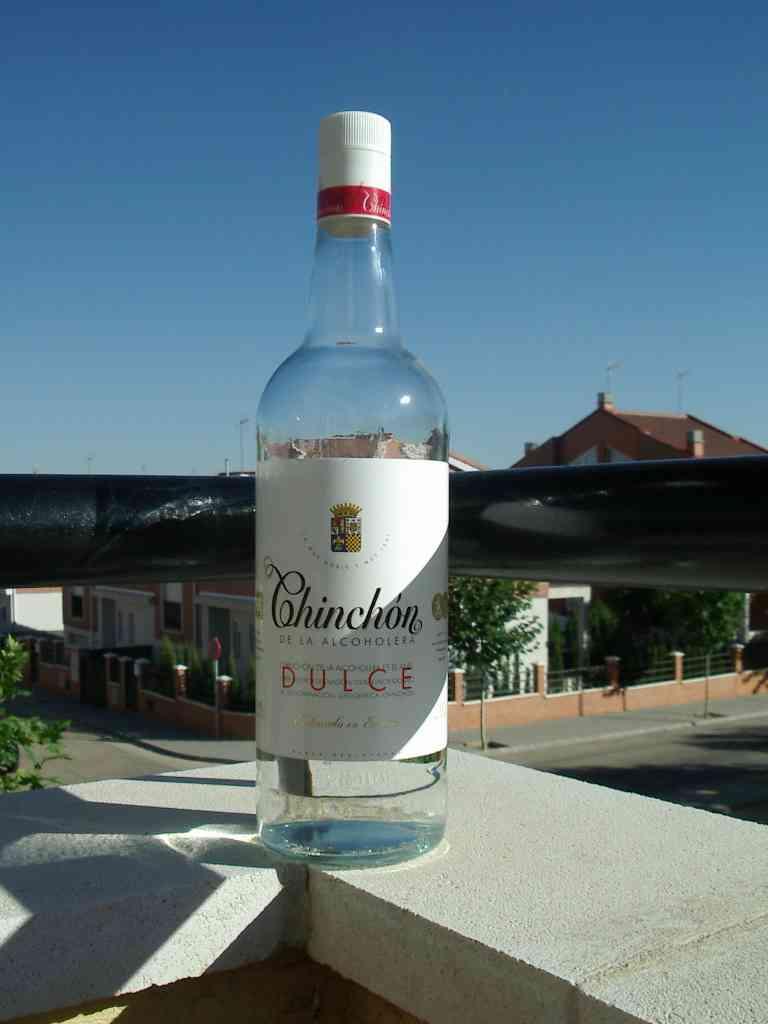} &
    \includegraphics[width=\linewidth,height=0.75\linewidth]{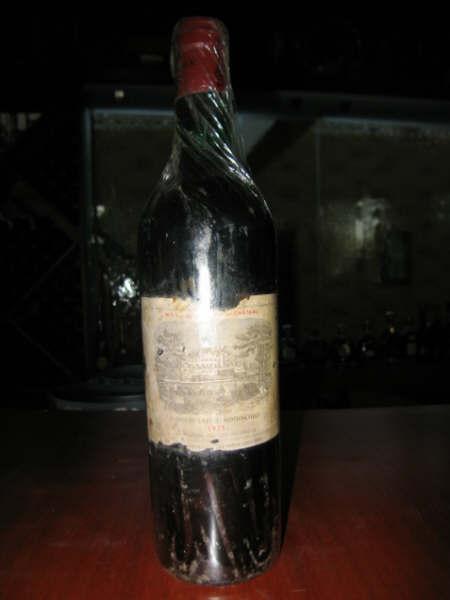}\\
    
    \footnotesize{\fontfamily{qhv}\selectfont \textbf{GT: } Ouzo} \par      {\color{blue}\footnotesize{\fontfamily{qhv}\selectfont \textbf{Ouzo: 0.53}}}
    \par {\footnotesize{\fontfamily{qhv}\selectfont {Vodka: 0.14}}}
    \par {\footnotesize{\fontfamily{qhv}\selectfont {BirchB: 0.04}}}
    &
    \footnotesize{\fontfamily{qhv}\selectfont \textbf{GT: } RootB} \par {\color{blue}\footnotesize{\fontfamily{qhv}\selectfont \textbf{RootB: 0.74}}}
    \par {\footnotesize{\fontfamily{qhv}\selectfont {QuinW: 0.07}}}
    \par {\footnotesize{\fontfamily{qhv}\selectfont {Vodka: 0.02}}}
    &
    \footnotesize{\fontfamily{qhv}\selectfont \textbf{GT: } Sarsap} \par {\color{blue}\footnotesize{\fontfamily{qhv}\selectfont \textbf{Sarsap: 0.97}}}
    \par {\footnotesize{\fontfamily{qhv}\selectfont {QW: 3.4e-5}}}
    \par {\footnotesize{\fontfamily{qhv}\selectfont {Bitter: 3.0e-5}}}
    &
     \footnotesize{\fontfamily{qhv}\selectfont \textbf{GT: } Vodka} \par {\color{blue}\footnotesize{\fontfamily{qhv}\selectfont \textbf{Vodka: 0.58}}}
    \par {\footnotesize{\fontfamily{qhv}\selectfont {Ouzo:0.15}}}
    \par {\footnotesize{\fontfamily{qhv}\selectfont {Pepsi:0.13}}}
    &
    \footnotesize{\fontfamily{qhv}\selectfont \textbf{GT: } Biiter} \par      {\color{blue}\footnotesize{\fontfamily{qhv}\selectfont \textbf{Bitter: 0.31}}}
    \par {\footnotesize{\fontfamily{qhv}\selectfont {BirchB: 0.17}}}
    \par {\footnotesize{\fontfamily{qhv}\selectfont {RootB: 0.0}}}
    &
    \footnotesize{\fontfamily{qhv}\selectfont \textbf{GT: } Guinn} \par {\color{blue}\footnotesize{\fontfamily{qhv}\selectfont \textbf{Guinn: 0.40}}}
    \par {\footnotesize{\fontfamily{qhv}\selectfont {Drambui: 0.15}}}
    \par {\footnotesize{\fontfamily{qhv}\selectfont {Sauterne: 0.08}}}
    &
    \footnotesize{\fontfamily{qhv}\selectfont \textbf{GT: } Ouzo} \par {\color{red}\footnotesize{\fontfamily{qhv}\selectfont \textbf{Vodka: 0.48}}}
    \par {\footnotesize{\fontfamily{qhv}\selectfont {Ouzo: 0.18}}}
    \par {\footnotesize{\fontfamily{qhv}\selectfont {Sauterne: 0.14}}}
    &
     \footnotesize{\fontfamily{qhv}\selectfont \textbf{GT: } Ouzo} \par {\color{red}\footnotesize{\fontfamily{qhv}\selectfont \textbf{Sauterne: 0.90}}}
    \par {\footnotesize{\fontfamily{qhv}\selectfont {Rootb: 3.0e-2}}}
    \par {\footnotesize{\fontfamily{qhv}\selectfont {Chablis: 2.2e-2}}}
    \\
    & \\
    \includegraphics[width=\linewidth,height=0.75\linewidth]{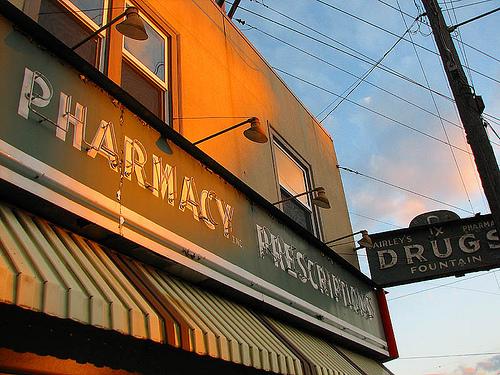} 
    & 
    \includegraphics[width=\linewidth,height=0.75\linewidth]{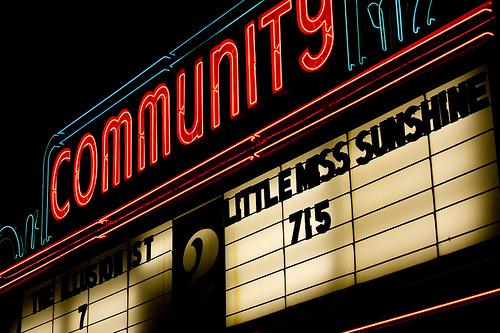}
    & 
    \includegraphics[width=\linewidth,height=0.75\linewidth]{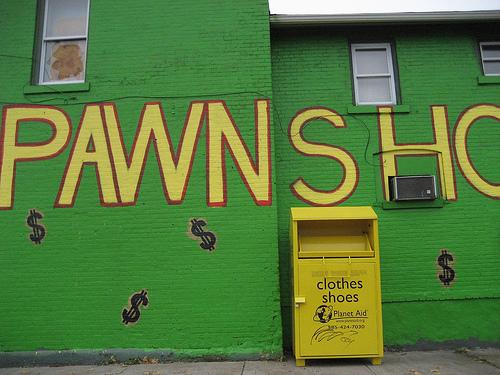}
    & 
    \includegraphics[width=\linewidth,height=0.75\linewidth]{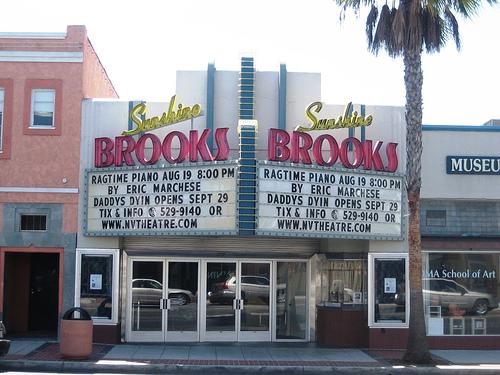}
    &
    \includegraphics[width=\linewidth,height=0.75\linewidth]{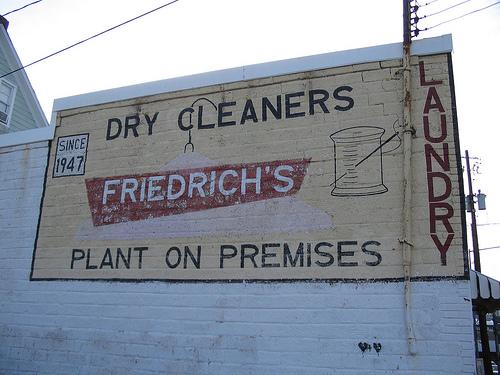} 
    & 
    \includegraphics[width=\linewidth,height=0.75\linewidth]{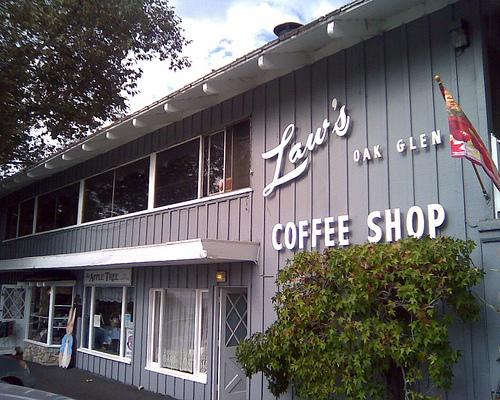}
    & 
    \includegraphics[width=\linewidth,height=0.75\linewidth]{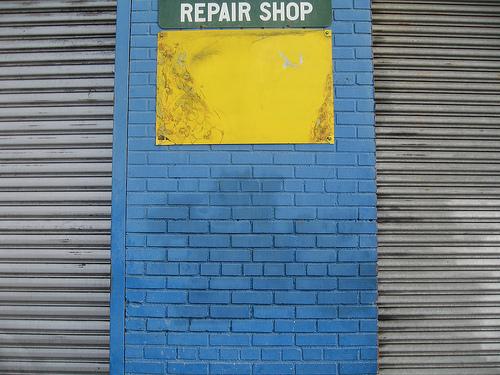}
    & 
    \includegraphics[width=\linewidth,height=0.75\linewidth]{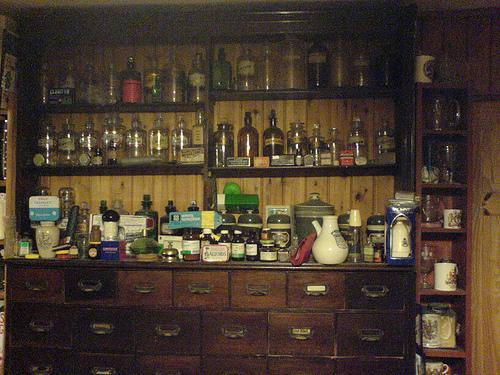}
    \\
     \footnotesize{\fontfamily{qhv}\selectfont \textbf{GT: } Pharma} \par {\color{blue}\footnotesize{\fontfamily{qhv}\selectfont \textbf{Pharma: 0.42}}}
    \par {\footnotesize{\fontfamily{qhv}\selectfont {Funeral: 0.13}}}
    \par {\footnotesize{\fontfamily{qhv}\selectfont {Cafe: 0.08}}}
    &
     \footnotesize{\fontfamily{qhv}\selectfont \textbf{GT: } Theatre} \par {\color{blue}\footnotesize{\fontfamily{qhv}\selectfont \textbf{Theatre: 0.83}}}
    \par {\footnotesize{\fontfamily{qhv}\selectfont {Diner: 0.01}}}
    \par {\footnotesize{\fontfamily{qhv}\selectfont {Pharma: 0.01}}}
    &
     \footnotesize{\fontfamily{qhv}\selectfont \textbf{GT: } PawnS} \par {\color{blue}\footnotesize{\fontfamily{qhv}\selectfont \textbf{PawnS: 0.38}}}
    \par {\footnotesize{\fontfamily{qhv}\selectfont {School: 0.27}}}
    \par {\footnotesize{\fontfamily{qhv}\selectfont {MedicalC: 0.16}}}
    &
     \footnotesize{\fontfamily{qhv}\selectfont \textbf{GT: } Theatre} \par {\color{blue}\footnotesize{\fontfamily{qhv}\selectfont \textbf{Theatre: 0.99}}}
    \par {\footnotesize{\fontfamily{qhv}\selectfont {BookS: 4.8e-3}}}
    \par {\footnotesize{\fontfamily{qhv}\selectfont {Disc.H: 9.5e-4}}}
    &
    \footnotesize{\fontfamily{qhv}\selectfont \textbf{GT: } DryCl} \par {\color{blue}\footnotesize{\fontfamily{qhv}\selectfont \textbf{DryCl:  0.20}}}
    \par {\footnotesize{\fontfamily{qhv}\selectfont {Resta: 0.11}}}
    \par {\footnotesize{\fontfamily{qhv}\selectfont {TeaH:  0.09}}}
    &
     \footnotesize{\fontfamily{qhv}\selectfont \textbf{GT: } Cafe} \par {\color{blue}\footnotesize{\fontfamily{qhv}\selectfont \textbf{Cafe: 0.46}}}
    \par {\footnotesize{\fontfamily{qhv}\selectfont {Resta: 0.16}}}
    \par {\footnotesize{\fontfamily{qhv}\selectfont {Barber: 0.07}}}
    &
     \footnotesize{\fontfamily{qhv}\selectfont \textbf{GT: } RepairS} \par {\color{red}\footnotesize{\fontfamily{qhv}\selectfont \textbf{Barber: 0.62}}}
    \par {\footnotesize{\fontfamily{qhv}\selectfont {RepairS: 0.09}}}
    \par {\footnotesize{\fontfamily{qhv}\selectfont {School: 8.3e-3}}}
    &
     \footnotesize{\fontfamily{qhv}\selectfont \textbf{GT: } Pharma} \par {\color{red}\footnotesize{\fontfamily{qhv}\selectfont \textbf{Tobacco: 0.37}}}
    \par {\footnotesize{\fontfamily{qhv}\selectfont {Pharma: 0.33}}}
    \par {\footnotesize{\fontfamily{qhv}\selectfont {Bistro: 0.06}}}

\end{tabular}
\end{center}
\caption{Classification results. The top-3 probabilities of a given image assigned by the output our model is shown along the Ground Truth. Notice that without reading, the classification task is impossible to perform even for humans. Blue and red are used to display correct and incorrect predictions respectively.}
\label{fig:long}
\end{figure*}
Secondly, by combining the explored fusion methods along with Fisher Vectors did not provide a significant advantage. 
A straightforward concatenation operation between the FV and the visual features reinforces the notion that both modalities contain discriminative and orthogonal features well suited for this task. As an additional advantage, by employing concatenation the model convergences faster while at the same time providing a better performance.

\subsection{Qualitative Results}

Fine-grained classification probabilities obtained from our model output are depicted in Figure~\ref{fig:long}. The textual features employed are able to generalize to unseen textual instances or named entities such as the case of bottle brands or business places. 
We can observe that our model has a hard time reading handwritten text or vertical textual occurrences, thus wrongly predicting a class, such as the example shown at the first row, seventh column. Nonetheless, the model seems to be capturing text morphology, as can be seen on the prediction of the class 'pawn shop'. 
Finally on the last two samples on each row, there are not enough guiding textual features and the model relies only on similar visual features. Nonetheless, classifying these samples correctly are a hard task even for humans.

\subsection{Fine-grained Image Retrieval}

In the same manner as the work presented in~\cite{karaoglu2017words} and~\cite{bai2018integrating}, we conduct a retrieval experiment by utilizing the computed vector of the last output layer of the proposed model as retrieval features. 
\begin{table}[H]
\begin{center}
\begin{tabular}{l|ll}
             \textbf{Method} & \textbf{Con-Text} & \textbf{Drink Bottle} \\ \hline
Bai*\cite{bai2018integrating}  & $62.87$                & $60.80$                 \\
Ours & \boldmath$64.52$            & \boldmath$62.91$                
\end{tabular}
\end{center}
\caption{Retrieval results on the evaluated datasets. The results on Con-Text are based on our implementation of the method by~\cite{bai2018integrating} since there is no publicly available code. The retrieval scores are depicted in terms of the mAP(\%).}
\label{tab:retrieval_results}
\end{table}
We take the approach of query by example, that is, given a sample image that belongs to a specific class, the system must return a ranked list of similar classes as the query. The metric employed to conduct this experiment is the cosine similarity.
The proposed method is more robust at the moment of employing a combination of visual and textual features which are discriminative enough to conduct a different task successfully as it is the case in fine-grained image retrieval. The retrieval quantitative performance for both datasets is shown in Table~\ref{tab:retrieval_results}, for qualitative results please refer to the Supplementary Material.
\section{Conclusions and Future Work}
\label{sec:Conclusions}

In this work, we have presented a deep neural network framework suitable for a fine-grained classification task. 
Through extensive experiments conducted, we have presented that leveraging textual information is a key approach to extract information from images. Exploiting these textual cues can pave the road towards more holistic computer vision models of scene understanding. We have shown that current text recognizers that are limited by a dictionary are not the best alternative for this task, because it requires a recognizer able to generalize out of vocabulary words from unseen samples. 
Additionally, we have analyzed the fact that using semantic embeddings in a fine-grained classification task do not produce the best results due to the related semantic space shared across similar classes. By integrating state-of-the-art techniques and constructing a powerful morphological descriptor from text contained in images, we show that a better suited feature for this task can be learned. Such a feature proves to be useful for a fine-grained classification task as well as for query-by-example image retrieval. Leveraging this robust textual feature yields state-of-the-art results in both tasks across the assessed datasets. Classification and retrieval is possible due to the discriminative features learnt by the model.
As future work, we plan to develop a morphological descriptor that captures the same discriminative features using a smaller feature dimension. A continuous valued embedding can replace the binary PHOC while preserving the generalization ability of unseen samples. We want to explore the usefulness of this embedding in other computer vision tasks such as visual question answering~\cite{biten2019stvqa, singh2019towards} and text-based image retrieval. 

{\small
\bibliographystyle{ieee}
\bibliography{biblio.bib}
}

\end{document}